\PassOptionsToPackage{unicode}{hyperref}
\PassOptionsToPackage{hyphens}{url}
\PassOptionsToPackage{dvipsnames,svgnames,x11names}{xcolor}
\documentclass[
]{article}

\usepackage{amsmath,amssymb}
\usepackage{iftex}
\ifPDFTeX
  \usepackage[T1]{fontenc}
  \usepackage[utf8]{inputenc}
  \usepackage{textcomp} 
\else 
  \usepackage{unicode-math}
  \defaultfontfeatures{Scale=MatchLowercase}
  \defaultfontfeatures[\rmfamily]{Ligatures=TeX,Scale=1}
\fi
\usepackage{lmodern}
\ifPDFTeX\else  
\fi
\IfFileExists{upquote.sty}{\usepackage{upquote}}{}
\IfFileExists{microtype.sty}{
  \usepackage[]{microtype}
  \UseMicrotypeSet[protrusion]{basicmath} 
}{}
\makeatletter
\@ifundefined{KOMAClassName}{
  \IfFileExists{parskip.sty}{%
    \usepackage{parskip}
  }{
    \setlength{\parindent}{0pt}
    \setlength{\parskip}{6pt plus 2pt minus 1pt}}
}{
  \KOMAoptions{parskip=half}}
\makeatother
\usepackage{xcolor}
\ifx\paragraph\undefined\else
  \let\oldparagraph\paragraph
  \renewcommand{\paragraph}[1]{\oldparagraph{#1}\mbox{}}
\fi
\ifx\subparagraph\undefined\else
  \let\oldsubparagraph\subparagraph
  \renewcommand{\subparagraph}[1]{\oldsubparagraph{#1}\mbox{}}
\fi

\usepackage{longtable,booktabs,array}
\usepackage{calc} 
\usepackage{etoolbox}
\makeatletter
\patchcmd\longtable{\par}{\if@noskipsec\mbox{}\fi\par}{}{}
\makeatother
\IfFileExists{footnotehyper.sty}{\usepackage{footnotehyper}}{\usepackage{footnote}}
\makesavenoteenv{longtable}
\usepackage{graphicx}
\makeatletter
\def\maxwidth{\ifdim\Gin@nat@width>\linewidth\linewidth\else\Gin@nat@width\fi}
\def\maxheight{\ifdim\Gin@nat@height>\textheight\textheight\else\Gin@nat@height\fi}
\makeatother
\setkeys{Gin}{width=\maxwidth,height=\maxheight,keepaspectratio}
\makeatletter
\def\fps@figure{htbp}
\makeatother
\NewDocumentCommand\citeproctext{}{}

\makeatletter
 \let\@cite@ofmt\@firstofone
 \def\@biblabel#1{}
 \def\@cite#1#2{{#1\if@tempswa , #2\fi}}
\makeatother
\newlength{\cslhangindent}
\newlength{\csllabelwidth}
\newenvironment{CSLReferences}[2] 
 {\begin{list}{}{%
  \setlength{\itemindent}{0pt}
  \setlength{\leftmargin}{0pt}
  \setlength{\parsep}{0pt}
  \ifodd #1
   \setlength{\leftmargin}{\cslhangindent}
   \setlength{\itemindent}{-1\cslhangindent}
  \fi
  \setlength{\itemsep}{#2\baselineskip}}}
 {\end{list}}
\usepackage{calc}

\newcommand{\CSLLeftMargin}[1]{\parbox[t]{\csllabelwidth}{\strut#1\strut}}
\newcommand{\CSLRightInline}[1]{\parbox[t]{\linewidth - \csllabelwidth}{\strut#1\strut}}

\usepackage{arxiv}
\usepackage{orcidlink}
\usepackage{amsmath}
\usepackage[T1]{fontenc}
\makeatletter
\@ifpackageloaded{caption}{}{\usepackage{caption}}
\AtBeginDocument{%
\ifdefined\contentsname
  \renewcommand*\contentsname{Table of contents}
\else
  \newcommand\contentsname{Table of contents}
\fi
\ifdefined\listfigurename
  \renewcommand*\listfigurename{List of Figures}
\else
  \newcommand\listfigurename{List of Figures}
\fi
\ifdefined\listtablename
  \renewcommand*\listtablename{List of Tables}
\else
  \newcommand\listtablename{List of Tables}
\fi
\ifdefined\figurename
  \renewcommand*\figurename{Figure}
\else
  \newcommand\figurename{Figure}
\fi
\ifdefined\tablename
  \renewcommand*\tablename{Table}
\else
  \newcommand\tablename{Table}
\fi
}
\@ifpackageloaded{float}{}{\usepackage{float}}
\floatstyle{ruled}
\@ifundefined{c@chapter}{\newfloat{codelisting}{h}{lop}}{\newfloat{codelisting}{h}{lop}[chapter]}
\floatname{codelisting}{Listing}

\makeatother
\makeatletter
\@ifpackageloaded{caption}{}{\usepackage{caption}}
\@ifpackageloaded{subcaption}{}{\usepackage{subcaption}}
\makeatother
\ifLuaTeX
  \usepackage{selnolig}  
\fi
\usepackage{bookmark}

\IfFileExists{xurl.sty}{\usepackage{xurl}}{} 
\hypersetup{
  pdftitle={Power-scaled Bayesian inference with score-based generative models},
  pdfauthor={Huseyin Tuna Erdinc; Yunlin Zeng; Abhinav Prakash Gahlot; Felix J. Herrmann},
  colorlinks=true,
  linkcolor={blue},
  filecolor={Maroon},
  citecolor={Blue},
  urlcolor={Blue},
  pdfcreator={LaTeX via pandoc}}

\title{Power-scaled Bayesian inference with score-based generative
models}
\def\asep{\\\\\\ } 
\def\asep{\And }
\author{\textbf{Huseyin Tuna Erdinc}\\\\Georgia Institute of
Technology\\\\\asep\textbf{Yunlin Zeng}\\\\Georgia Institute of
Technology\\\\\asep\textbf{Abhinav Prakash Gahlot}\\\\Georgia Institute
of Technology\\\\\asep\textbf{Felix J. Herrmann}\\\\Georgia Institute of
Technology\\\\}
\date{}
\begin{document}
\maketitle
\begin{abstract}
We propose a score-based generative algorithm for sampling from
power-scaled priors and likelihoods within the Bayesian inference
framework. Our algorithm enables flexible control over prior--likelihood
influence without requiring retraining for different power-scaling
configurations. Specifically, we focus on synthesizing seismic velocity
models conditioned on imaged seismic. Our method enables sensitivity
analysis by sampling from intermediate power posteriors, allowing us to
assess the relative influence of the prior and likelihood on samples of
the posterior distribution. Through a comprehensive set of experiments,
we evaluate the effects of varying the power parameter in different
settings: applying it solely to the prior, to the likelihood of a
Bayesian formulation, and to both simultaneously. The results show that
increasing the power of the likelihood up to a certain threshold
improves the fidelity of posterior samples to the conditioning data
(e.g., seismic images), while decreasing the prior power promotes
greater structural diversity among samples. Moreover, we find that
moderate scaling of the likelihood leads to a reduced shot data
residual, confirming its utility in posterior refinement.
\end{abstract}

\newcommand{\argmin}{\mathop{\mathrm{argmin}\,}\limits}
\newcommand{\argmax}{\mathop{\mathrm{argmax}\,}\limits}

\[
\def\textsc#1{\dosc#1\csod} 
\def\dosc#1#2\csod{{\rm #1{\small #2}}} 
\]

\section{Introduction}\label{introduction}

Subsurface velocity model generation forms a critical component of
hydrocarbon exploration {[}1{]}, subsurface monitoring {[}2{]}, and
numerous other geophysical applications {[}3{]}. Typically, subsurface
characterization is achieved by analyzing the Earth's response to
physical stimuli, such as electrodynamics, gravity, and acoustic wave
propagation, to variations in subsurface properties. The resulting
tomographic measurements are processed into images for downstream
interpretation. In this work, we focus specifically on modeling acoustic
properties by probing the Earth's interior using acoustic waves.
However, the proposed methodology is not limited to this particular
application and can be extended to a broad class of inverse problems.

Among the various inversion techniques, Full-Waveform Inversion (FWI)
has emerged as a leading method due to its ability to resolve
high-resolution acoustic models in complex geological settings {[}4{]}.
Despite its strengths, FWI suffers from several practical limitations:
it is computationally intensive, prone to convergence to local minima,
and sensitive to initial models due to the problem's inherently
nonlinear and ill-posed nature {[}5{]}. Additionally, it requires
repeated solutions of wave-equation-based partial differential equations
(PDEs), which significantly increases computational cost.

To alleviate these challenges, recent research has explored generative
models that leverage physics-informed summary statistics---such as
common-image gathers (CIGs) {[}6{]}, {[}7{]}, {[}8{]}, {[}9{]} or
reverse-time migration (RTM) images {[}10{]}, {[}11{]}---to guide the
inversion process. While promising, a common criticism of these
generative approaches is that they often rely heavily on strong,
structured priors, which can dominate the inference and limit the
model's responsiveness to observed data. In contrast, methods such as in
{[}12{]} employ highly non-informative priors, placing more emphasis on
the observed data. However, this comes at the cost of reduced
regularization, which can affect robustness---especially in
underdetermined or noisy regimes---and may increase susceptibility to
local minima during inference.

Building on these insights, we propose a novel framework that combines
classical Bayesian inference with score-based generative models trained
on geological structure-consistent priors. We introduce a modification
to the score-based sampling process that enables sampling from the
power-scaled versions of the prior and likelihood, allowing explicit
control over their relative influence during the inference process. We
validate our approach using RTM images as physics-based summary
statistics {[}13{]}, {[}14{]} produced using a smoothed background model
{[}6{]} and demonstrate its performance in generating diverse and
data-consistent subsurface velocity models. It is important to note
that, although the background models used for RTM are smoothed, they are
not kinematically incorrect. As a result, the RTM images preserve key
information from the original shot data. While prior work such as
{[}12{]}, {[}15{]} directly utilizes shot records for inference, our
method leverages RTM images.

\section{Theoretical explanation}\label{theoretical-explanation}

\subsection{Seismic imaging and Bayesian
inference}\label{seismic-imaging-and-bayesian-inference}

Estimation of the unknown subsurface property such as the acoustic
wavespeed, \(\mathbf{x}\), requires solving an inverse problem using
observed data \(\mathbf{y}\) . In our context, we can define our forward
problem as:

\begin{equation}\phantomsection\label{eq-forward_problem}{
  \mathbf{y} = \mathcal{F}(\boldsymbol{x}) + \boldsymbol{\epsilon}, \qquad \boldsymbol{\epsilon} \sim p(\boldsymbol{\epsilon})
}\end{equation}

where \(\mathcal{F}\) represents the nonlinear forward operator and
\(\boldsymbol{\epsilon}\) is bandlimited noise. The main complexity of
inverting this problem stems from the nontrivial null-space of the
forward operator and the compounding effect of the noise {[}4{]}. As a
result, multiple velocity models can explain the observed data equally
well, which necessitates the use of a Bayesian framework to properly
quantify uncertainties. Bayesian inference provides a probabilistic
formulation for inverse problems by computing the posterior probability
density function (pdf) using Bayes' rule:

\begin{equation}\phantomsection\label{eq-bayes}{
p(\mathbf{x}|\mathbf{y}) = \frac{p(\mathbf{y}|\mathbf{x}) p(\mathbf{x})}{p(\mathbf{y})},
}\end{equation}

where \(p(\mathbf{x})\) is the prior that describes available
information about the velocity \(\mathbf{x}\) before the inference
process, and \(p(\mathbf{y}|\mathbf{x})\) is the likelihood function,
which, given any model value \(\mathbf{x}\) calculates the probability
of observing the imaged data---denoted, with a slight abuse of notation,
also by \(\mathbf{y}\)---given any model \(\mathbf{x}\). The likelihood
is used to describe how well \(\mathbf{y}\) matches the image generated
by a particular model \(\mathbf{x}\). The denominator,
\(p(\mathbf{y})\), is the evidence or marginal likelihood, serving as a
normalization constant to ensure that the posterior is a valid
probability distribution.

\subsection{Simulation-based inference via conditional score-based
networks}\label{simulation-based-inference-via-conditional-score-based-networks}

\hspace{0pt}Simulation-based inference (SBI) is a framework that allows
the training of surrogates for posterior pdf using neural estimators
{[}16{]}. The key idea is to use numerical simulators to generate
training pairs
\(\mathcal{D} = \{ (\mathbf{x}^{i}, \mathbf{y}^{i}) \}_{i=1}^{N}\),
where each pair consists of a set of subsurface properties
\(\mathbf{x}^{i}\) and the corresponding simulated observation
\(\mathbf{y}^{i}\) derived using the forward simulation. In this study,
rather than working directly with raw seismic data, we extract
reverse-time migration (RTM) images as summary statistics, which are
used as \(\mathbf{y}\) during both training and inference. This approach
retains key structural information while reducing the dimensionality and
complexity of the data. The resulting pairs
\((\mathbf{x}^{i}, \mathbf{y}^{i})\) are then used to train a
conditional generative network, which learns the posterior distribution
of the velocities conditioned on RTM images. In this study, we will use
conditional score-based generative models in an SBI setting.

Score-based models are density estimators that learn the annealed score
of the target distribution,
\(\nabla_{\mathbf{x}_{t}} \log p(\mathbf{x}_{t})\), where the annealed
distribution is defined as
\(p(\mathbf{x}_{t}) = p(\mathbf{x}) * \mathcal{N}(0, \sigma(t)^2 \boldsymbol{I})\).
Here, \(t\) denotes the time, and \(\sigma(t)\) is the time-dependent
noise schedule. The annealed distribution can also be interpreted as a
gradual corruption of the target distribution through the progressive
addition of Gaussian noise, a process that corresponds to diffusion.
Once the score function is completed, samples from the target
distribution can be generated by solving the stochastic differential
equation (SDE):

\begin{equation}\phantomsection\label{eq-SDE}{
\mathbf{x} = -(\dot{\sigma}(t) + \beta(t)\sigma(t))\sigma(t)
\nabla_{\mathbf{x}_{t}} \log p(\mathbf{x}_{t}) dt 
+ \sqrt{2\beta(t)} \sigma(t) d\boldsymbol{\omega}_t,
}\end{equation}

following existing strategies in {[}17{]}. In this expression,
\(\boldsymbol{\omega}_t\) is the standard Wiener process, and
\(\beta(t)\) is a function that describes the amount of stochastic noise
during the sampling process. If \(\beta(t) = 0\) for all t, then the
process becomes a probabilistic ordinary differential equation (ODE),
otherwise it represents time-varying Langevin diffusion SDE.

As proposed in {[}17{]}, we adopt a simplified score-learning objective
with \(\sigma(t) = t\), where \(\sigma\) is sampled directly from a
data-dependent log-normal distribution. The training objective can be
written as:

\begin{equation}\phantomsection\label{eq-original_loss}{
\boldsymbol{\widehat \theta} = \mathop{\mathrm{argmin}\,}\limits_{\boldsymbol{\theta}} \mathbb{E}_{\mathbf{x}\sim p(\mathbf{x})}
\mathbb{E}_{\sigma \sim \text{LogNormal}(P_{\text{mean}}, P^2_{\text{std}})}
\mathbb{E}_{\mathbf{n_{\sigma}}\sim \mathcal{N}(0,\sigma^2 \boldsymbol{I})}
\| D_{\boldsymbol{\theta}}(\mathbf{{x}} + \mathbf{n_{\sigma}};\sigma)-\mathbf{x} \|^2_2,
}\end{equation}

where the denoising network
\(D_{\boldsymbol{\theta}}(\mathbf{x} + \mathbf{n}_{\sigma}; \sigma)\),
with learnable parameters \(\boldsymbol{\theta}\), is trained to recover
the original image from a noisy input. \(P_{\text{mean}}\) and
\(P_{\text{std}}\) are dataset-dependent parameters that control the
log-normal noise schedule. The score function can then be estimated as
\(\nabla_{\mathbf{x}_t} \log p(\mathbf{x}_t) \approx \left( D_{\boldsymbol{\theta}}(\mathbf{x} + \mathbf{n}_{\sigma}; \sigma) - (\mathbf{x} + \mathbf{n}_{\sigma}) \right) / \sigma^2\).
This formulation corresponds to the unconditional case. To model
conditional distributions, we extend it to estimate the conditional
score
\(\nabla_{\mathbf{x}_{t}} \log p(\mathbf{x}_{t} \mid \mathbf{y})\). The
revised training objective for the network becomes:

\begin{equation}\phantomsection\label{eq-conditional_loss}{
\boldsymbol{\widehat \theta} = \mathop{\mathrm{argmin}\,}\limits_{\boldsymbol{\theta}} {\mathbb{E}}_{\mathbf{y}\sim p(\mathbf{y}\mid\mathbf{x})} 
\mathbb{E}_{\mathbf{x}\sim p(\mathbf{x})}
\mathbb{E}_{\sigma \sim \text{LogNormal}(P_{\text{mean}}, P^2_{\text{std}})}
\mathbb{E}_{\mathbf{n_{\sigma}}\sim \mathcal{N}(0,\sigma^2 \boldsymbol{I})}
\| D_{\boldsymbol{\theta}}(\mathbf{{x}} +\mathbf{n}_{\sigma},\mathbf{y};\sigma)-\mathbf{x} \|^2_2.
}\end{equation}

\subsection{Power-scaling in Bayesian
inference}\label{power-scaling-in-bayesian-inference}

Previous approaches to Bayesian inference in seismic inversion often
face a trade-off: either they rely heavily on strong priors, which can
diminish the influence of the observed data, or they place excessive
emphasis on the data, potentially at the expense of regularization. This
motivates the need for a principled mechanism to adjust the relative
influence of the prior and likelihood without incurring significant
computational costs.

To address this, we introduce power-scaling as a flexible tool to
modulate this balance. Power-scaling is a controlled,
distribution-agnostic technique for modifying probability distributions
by adjusting the relative influence of the likelihood and prior
{[}18{]}. Intuitively, it can be understood as a mechanism to either
amplify or attenuate the effect of each component in the posterior,
without requiring any specific parametric assumptions.

The power-scaled posterior is defined as:
\begin{equation}\phantomsection\label{eq-power_scaling}{
p^{\lambda,\alpha}(\mathbf{x}|\mathbf{y}) = \frac{p(\mathbf{y}|\mathbf{x})^{\lambda} p(\mathbf{x})^{\alpha}}{\int p(\mathbf{y}|\mathbf{x})^{\lambda} p(\mathbf{x})^{\alpha} \mathrm{d}\mathbf{x}}. 
}\end{equation}

Here, \(\lambda\) denotes the power-scaling factor applied to the
likelihood, and \(\alpha\) is the factor for the prior. From a
statistical perspective, power-scaling the likelihood (i.e., increasing
\(\lambda\)) mimics the effect of having more observations, thereby
concentrates the posterior around high-likelihood regions. Conversely,
reducing \(\lambda\) diminishes the influence of the data, resulting in
more diffuse posteriors. Similarly, scaling the prior by an exponent
\(\alpha\) adjusts its influence: values of \(\alpha > 1\) sharpen the
prior around its high-density regions and reinforce prior assumptions,
while smaller values of \(\alpha\) relax the prior and enable the
posterior to explore a broader range of plausible solutions.

In this study, we propose to estimate power-scaled posterior using
score-based generative modeling. The score of the log power-scaled
posterior can be written as:

\begin{equation}\phantomsection\label{eq-power_scaling_step1}{
\nabla_{\mathbf{{x}}}\log p^{\lambda, \alpha}(\mathbf{x}|\mathbf{y}) = \lambda \nabla_{\mathbf{{x}}}\log p(\mathbf{y}|\mathbf{x})+ \alpha  \nabla_{\mathbf{{x}}} \log p(\mathbf{x}).
}\end{equation}

A primary challenge in this formulation lies in estimating the gradient
of the likelihood term, which is often computationally intensive,
especially in high-dimensional and physics-based models such as seismic
inversion. To address this, we leverage a useful identity from Bayes'
rule that expresses the likelihood score in terms of the posterior and
prior scores:
\(\nabla_{\mathbf{{x}}} \log p(\mathbf{y}|\mathbf{x}) = \nabla_{\mathbf{{x}}} \log p(\mathbf{x}|\mathbf{y}) -  \nabla_{\mathbf{{x}}}\log p(\mathbf{x})\).
Substituting this into the power-scaled score expression yields the
following formulation:

\begin{equation}\phantomsection\label{eq-power_scaling_last_step}{
\nabla_{\mathbf{{x}}} \log p^{\lambda,\alpha}(\mathbf{x}|\mathbf{y}) = \lambda \nabla_{\mathbf{{x}}} \log p(\mathbf{x}|\mathbf{y})+ (\alpha - \lambda)  \nabla_{\mathbf{{x}}} \log p(\mathbf{x}).
}\end{equation}

This formulation corresponds to training surrogates for both the
posterior score,
\(\nabla_{\mathbf{x}} \log p(\mathbf{x} \mid \mathbf{y})\), and the
prior score, \(\nabla_{\mathbf{x}} \log p(\mathbf{x})\). Leveraging the
approach introduced in classifier-free guidance (CFG) {[}19{]}, we
estimate both scores using a single conditional network by randomly
masking the conditioning input during training. This eliminates the need
to train two separate networks. Once the score functions are learned, we
combine the posterior and prior scores at inference time using their
respective power coefficients, \(\lambda\) and \(\alpha\). To generate
valid samples from the power-scaled posterior, we incorporate a
Langevin-based correction step, such as predictor-corrector methods,
into the sampling loop to improve accuracy and stability {[}20{]},
{[}21{]}.

The primary advantage of our approach is its flexibility: only a single
amortized score network needs to be trained, yet it enables sampling
from a wide range of power-scaled posterior distributions by adjusting
power coefficients at inference time. This makes it ideal for
sensitivity analysis, robustness testing, or exploring different
prior-data tradeoffs without retraining. A key limitation, however, is
sampling efficiency. Unlike standard posterior sampling, our approach
requires Langevin-type correction steps at each sampling iteration,
which can increase computational cost---particularly when a large number
of steps are needed to ensure convergence. Nonetheless, the method still
avoids the need for expensive likelihood evaluations, and in practice,
we achieve efficient inference with a per-sample generation time of
approximately 4 seconds.

\section{Numerical case study}\label{numerical-case-study}

To evaluate the proposed methodology, we conduct a numerical case study
using a synthetic 3D Earth model derived from the Compass model, which
is representative of geological formations in the North Sea region
{[}22{]}. The training dataset is constructed by slicing 2D velocity
models from the 3D synthetic volume and pairing them with corresponding
reverse-time migration (RTM) images. The dataset consists of 800
training samples, each defined on a \(256\times 512\) grid with a
spatial resolution of \(12.5\) meters, covering an area of
\(3.2km\times 6.4\mathrm{km}\). Seismic data are simulated using \(16\)
sources and \(256\) receivers, with Ricker wavelets centered at
\(20\mathrm{Hz}\) dominant frequency and a recording duration of \(3.2\)
seconds. To simulate realistic acquisition conditions, \(10\mathrm{dB}\)
colored Gaussian noise is added to the shot records. RTM images are
generated using a Gaussian-smoothed 2D background velocity model. Both
wave simulation and migration are implemented using the open-source
seismic modeling and inversion package JUDI {[}23{]}.

To train the conditional score-based network for simultaneous posterior
and prior estimation, we adopt a classifier-free guidance (CFG)
strategy: the conditioning RTM input is randomly dropped with a
probability of \(0.2\) and replaced with Gaussian noise during training
{[}19{]}. For conditioning, RTM images are concatenated with the noisy
velocity samples, and the network, a U-Net architecture, is trained to
denoise and recover the clean velocity samples. The model is trained for
\(12\) GPU hours. For testing, we select an unseen RTM example that was
held out during training and perform posterior sampling using the
trained network.

\subsection{Prior scaling}\label{prior-scaling}

\begin{figure}

\includegraphics[width=1\textwidth,height=\textheight]{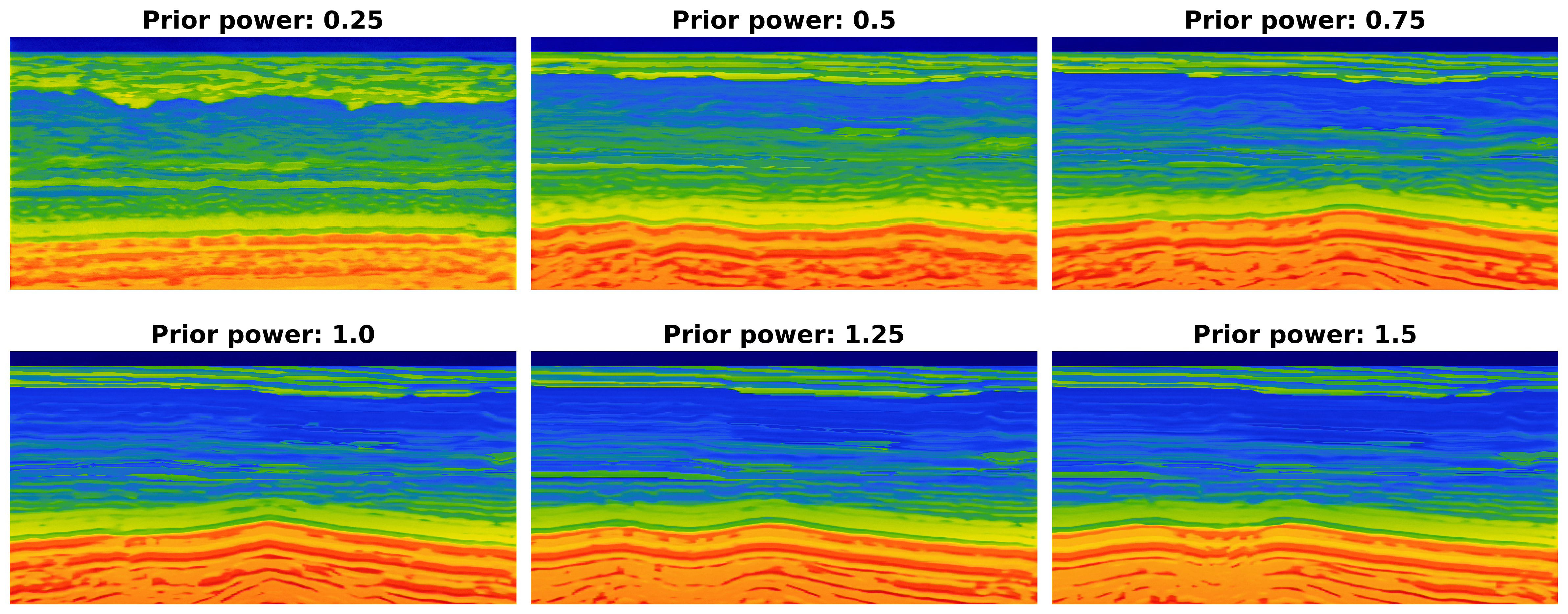}

\caption{\label{fig-progresssion_of_prior}Prior samples generated with
increasing power \(\alpha\) from \(0.25\) to \(1.5\). As \(\alpha\)
increases, the generated velocity models become progressively sharper
and more structurally consistent, especially in deeper layers. Lower
powers result in higher variability and less coherent geologic
features.}

\end{figure}

In the first experiment, we investigate the effect of power-scaling on
the learned prior distribution in isolation. Specifically, we set the
likelihood term aside and generate samples solely from the prior by
varying the coefficient in the power-scaled prior
\(p(\mathbf{x})^{\lambda}\). This allows us to analyze how different
levels of prior strength affect the structure and diversity of generated
velocity models. Figure~\ref{fig-progresssion_of_prior} shows the
evolution of prior samples as \(\alpha\) increases from \(0.25\) to
\(1.5\). To ensure a consistent comparison, all samples are generated
starting from the same random seed. At low prior powers (e.g.,
\(\alpha = 0.25\)), the generated samples exhibit high variability,
particularly in the deeper regions. Layer boundaries are less coherent
and geological structures appear diffuse and discontinuous. As
\(\alpha\) increases, the samples become increasingly regularized:
layers become more consistent and sharper. This reflects the fact that
the learning process concentrates probability mass around high-density
regions of the prior, which tend to correspond to well-structured
geological patterns seen in training data.

\subsection{Likelihood scaling in Bayesian
inference}\label{likelihood-scaling-in-bayesian-inference}

\begin{figure}

\includegraphics[width=1\textwidth,height=\textheight]{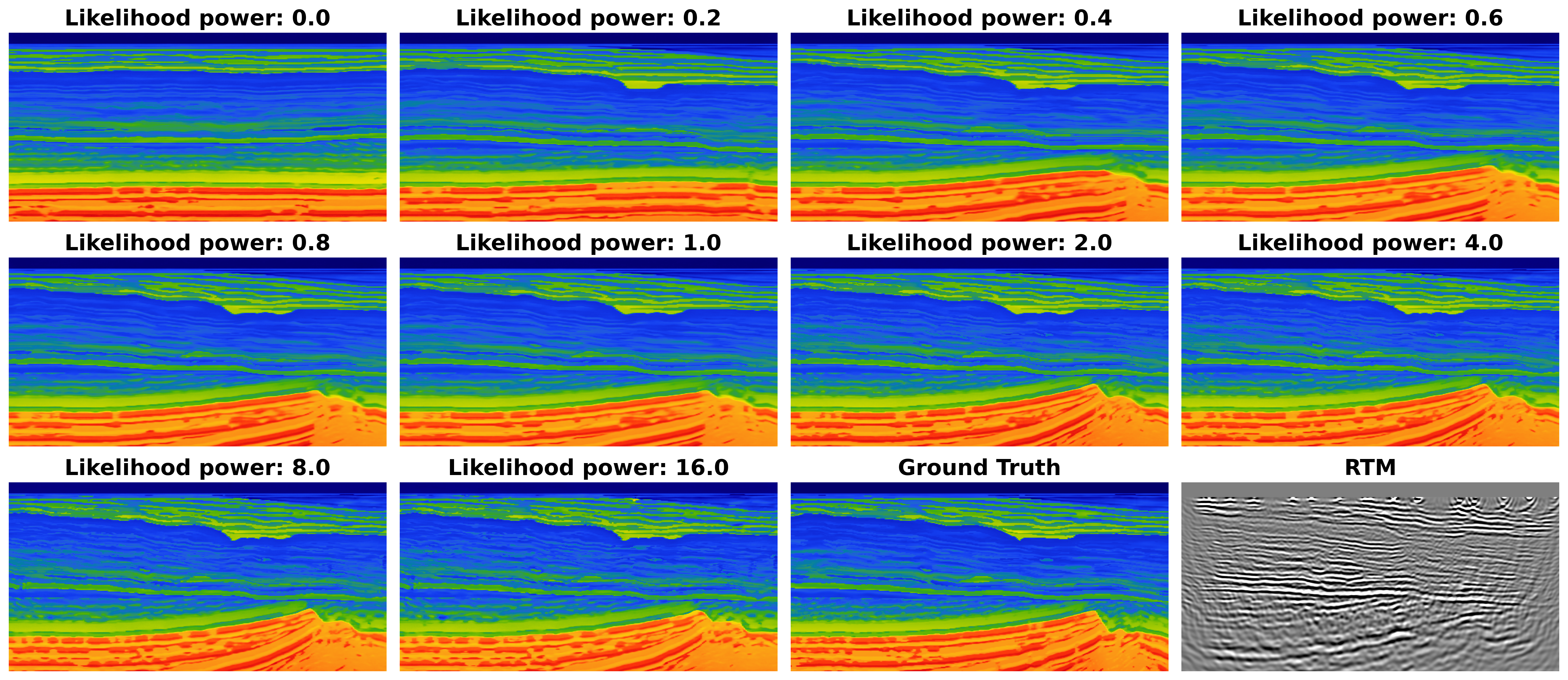}

\caption{\label{fig-progresssion_of_likelihood}Posterior samples
generated with varying likelihood power \(\lambda\) from \(0.0\) to
\(16.0\), with fixed prior power \(\alpha = 1\). As \(\lambda\)
increases, the samples incorporate more structure from the conditioning
RTM image. Maximum alignment with the ground truth occurs around
\(\lambda = 2.0\), beyond which overfitting and degradation in
performance become apparent.}

\end{figure}

\begin{figure}

\includegraphics[width=1\textwidth,height=\textheight]{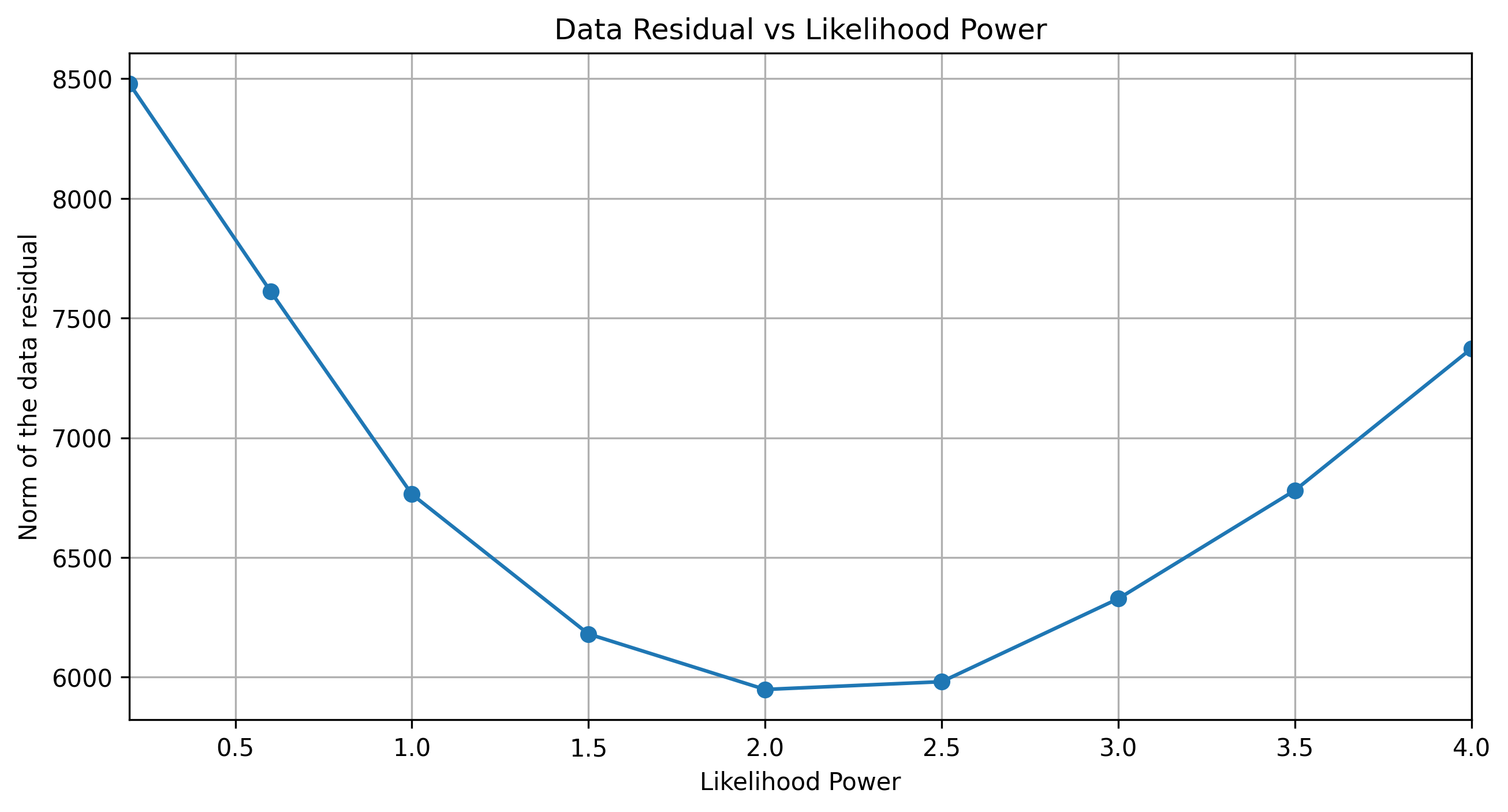}

\caption{\label{fig-residual}The \(\ell_2\)-norm of data residual as a
function of likelihood power \(\lambda\), with fixed prior power
\(\alpha = 1\). Residual decreases as \(\lambda\) increases, reaching a
minimum around \(\lambda = 2.0\), indicating optimal alignment with the
ground truth. Beyond this point, performance begins to degrade,
suggesting that excessive amplification of the likelihood leads to
overfitting or reduced generalization.}

\end{figure}

In this experiment, we fix the prior power at its standard Bayesian
setting, \(\alpha = 1\), and vary the likelihood power \(\lambda\) from
\(0\) to \(16\) for a specific RTM image. This allows us to investigate
how scaling the influence of the likelihood affects the resulting
posterior samples. Figure~\ref{fig-progresssion_of_likelihood} shows
velocity model samples generated with increasing values of \(\lambda\),
alongside the ground-truth velocity model and the RTM image used as the
conditioning input. At low values of \(\lambda\) (e.g.,
\(\lambda \leq 0.4\)), the contribution of the likelihood is minimal,
and the generated velocity models closely resemble unconditional prior
samples. The influence of the RTM conditioning is barely noticeable. As
\(\lambda\) increases, the posterior increasingly conforms to features
present in the RTM image. For instance, at \(\lambda=1.0\),
corresponding to standard Bayesian inference, the generated model shows
moderate data fidelity. Interestingly, we observe that data fidelity
continues to improve up to a certain point---most notably at
\(\lambda=2.0\)-after which performance begins to degrade. This behavior
is quantified in Figure~\ref{fig-residual}, which presents the data
residual (i.e., the difference between simulated shot data from the
generated velocity models and ground-truth shot data). The residual is
minimized at \(\lambda=2.0\), indicating optimal alignment with the
ground truth at this power level. This result highlights a key insight:
maximum posterior fidelity does not necessarily occur at the classical
Bayesian setting \(\lambda=1.0\), but rather at an upweighted likelihood
power. This demonstrates the practical value of power-scaling as a tool
for tuning the trade-off between prior regularization and data
conformity.

\subsection{Power-scaling compass}\label{power-scaling-compass}

\begin{figure}

\includegraphics[width=1\textwidth,height=\textheight]{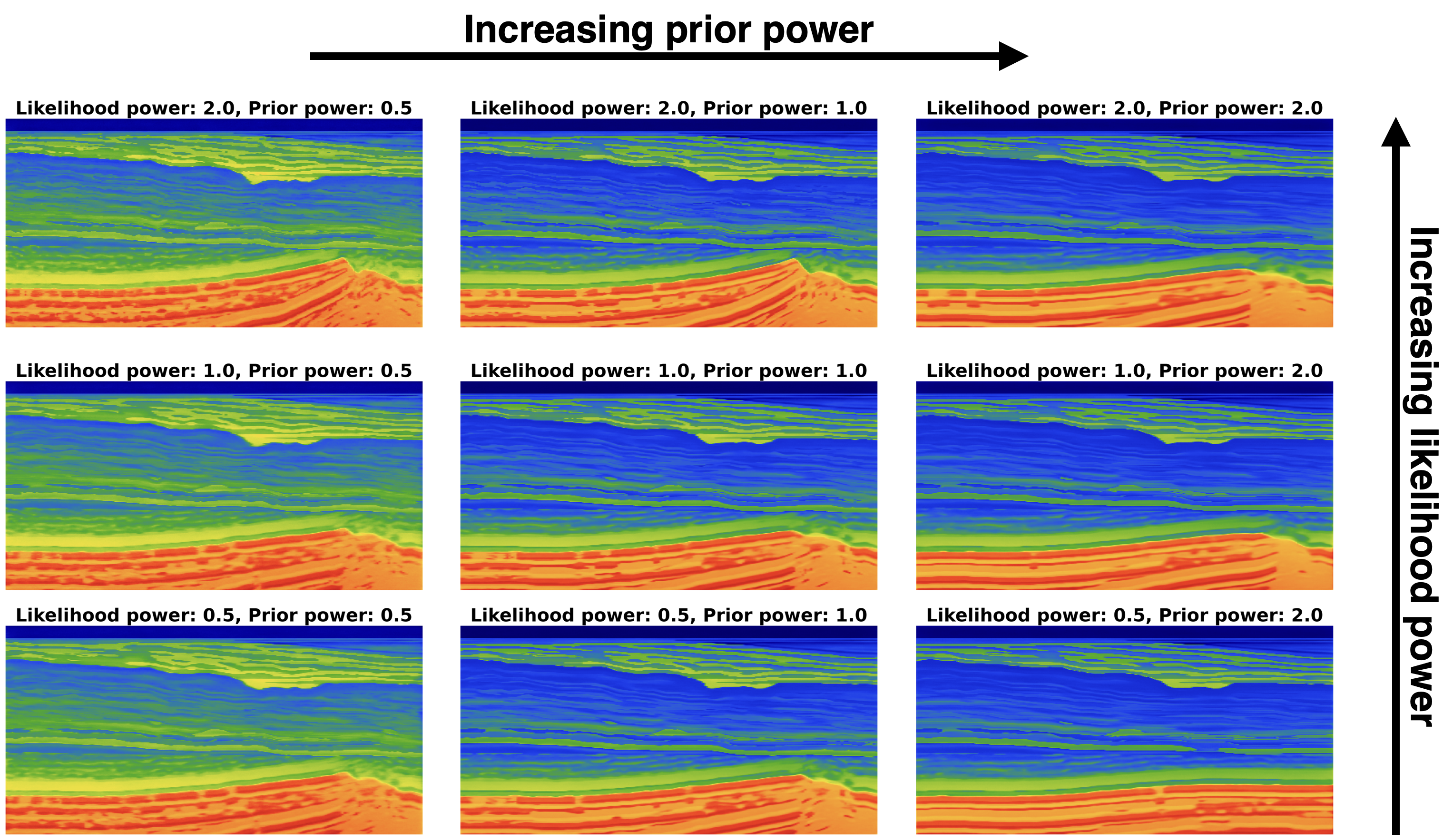}

\caption{\label{fig-power-scaling-compass}Posterior samples generated by
jointly varying likelihood power \(\lambda\) and prior power \(\alpha\).
Columns correspond to increasing prior power (left to right), while rows
correspond to increasing likelihood power (bottom to top). Samples with
larger \(\lambda\) better align with RTM conditioning data, while larger
\(\alpha\) values enforce structural consistency and sharper geological
layering. Lower prior powers increase generative diversity and loosen
geological constraints.}

\end{figure}

Having independently explored the effects of prior and likelihood power
scaling, we now investigate their joint influence by constructing a
power-scaling compass. In this setup, we simultaneously vary the prior
power \(\alpha\) and the likelihood power \(\lambda\) to study their
combined impact on posterior samples.
Figure~\ref{fig-power-scaling-compass} presents posterior samples
generated under various combinations of \(\alpha \in {0.5,1.0, 2.0}\)
and \(\lambda \in {0.5,1.0, 2.0}\). This grid-like layout allows us to
visually interpret how the balance between prior and likelihood power
influences the behavior of the samples.

We observe that when likelihood power exceeds prior power, the generated
samples exhibit greater alignment with the conditioning RTM
image---demonstrating stronger adherence to observed data. Conversely,
when prior power dominates, the samples exhibit smoother layer
transitions and more geologically consistent patterns, reflecting
stronger regularization from the prior distribution. Notably, reducing
the prior power relaxes the structural constraints in the model and
leads to increased variability and the emergence of less structured or
more exploratory features. This visualization highlights how
power-scaling enables fine-grained control over the trade-off between
data fidelity and prior-driven structural regularity, offering a
versatile tool for interpretability in seismic inversion tasks.

\section{Conclusions}\label{conclusions}

In this study, we proposed a method for applying power-scaling to
seismic velocity model generation within a Bayesian inference framework.
Our approach is built on a single amortized score-based generative model
trained using a classifier-free guidance strategy, enabling simultaneous
generation of both prior and posterior samples. Crucially, this design
allows flexible control over the relative influence of the prior and
likelihood without requiring expensive likelihood evaluations, as all
modifications are made at the inference stage. Experimental results
demonstrate that power-scaling offers a principled way to relax or
constrain the prior, making it a valuable tool for generating diverse
samples during training or for performing inference under less
influential priors. Moreover, increasing the likelihood power was shown
to improve data fidelity, enabling better alignment between posterior
samples and observed seismic features. In future work, we aim to conduct
a more rigorous investigation into the role of power-scaling in
uncertainty quantification and its impact on interpretability and
robustness in seismic inversion tasks.

\section{Acknowledgement}\label{acknowledgement}

This research was carried out with the support of Georgia Research
Alliance, partners of the ML4Seismic Center. During the preparation of
this work, the authors used ChatGPT to refine sentence structures and
improve readability. After using this service, the authors reviewed and
edited the content as needed and take full responsibility for the
content of the publication.

\section{References}\label{references}

\phantomsection\label{refs}


\begin{CSLReferences}{0}{0}
\bibitem[\citeproctext]{ref-zhang2022high}
\CSLLeftMargin{{[}1{]} }%
\CSLRightInline{L. Zhang, S. Sengupta, C. Parekh, and R.
Eliott-Lockhart, {``High-frequency FWI velocity model building for
hydrocarbon delineation,''} in \emph{SEG international exposition and
annual meeting}, SEG, 2022, p. D011S043R004.}

\bibitem[\citeproctext]{ref-gahlot2024uncertainty}
\CSLLeftMargin{{[}2{]} }%
\CSLRightInline{A. P. Gahlot, R. Orozco, Z. Yin, and F. J. Herrmann,
{``An uncertainty-aware digital shadow for underground multimodal CO2
storage monitoring,''} \emph{arXiv preprint arXiv:2410.01218}, 2024.}

\bibitem[\citeproctext]{ref-wagner2021overview}
\CSLLeftMargin{{[}3{]} }%
\CSLRightInline{F. M. Wagner and S. Uhlemann, {``An overview of
multimethod imaging approaches in environmental geophysics,''}
\emph{Advances in Geophysics}, vol. 62, pp. 1--72, 2021.}

\bibitem[\citeproctext]{ref-tarantola1984inversion}
\CSLLeftMargin{{[}4{]} }%
\CSLRightInline{A. Tarantola, {``Inversion of seismic reflection data in
the acoustic approximation,''} \emph{Geophysics}, vol. 49, no. 8, pp.
1259--1266, 1984.}

\bibitem[\citeproctext]{ref-virieux2009overview}
\CSLLeftMargin{{[}5{]} }%
\CSLRightInline{J. Virieux and S. Operto, {``An overview of
full-waveform inversion in exploration geophysics,''} \emph{Geophysics},
vol. 74, no. 6, pp. WCC1--WCC26, 2009.}

\bibitem[\citeproctext]{ref-yin2024wise}
\CSLLeftMargin{{[}6{]} }%
\CSLRightInline{Z. Yin, R. Orozco, M. Louboutin, and F. J. Herrmann,
{``WISE: Full-waveform variational inference via subsurface
extensions,''} \emph{Geophysics}, vol. 89, no. 4, pp. A23--A28, 2024.}

\bibitem[\citeproctext]{ref-yin2025wiser}
\CSLLeftMargin{{[}7{]} }%
\CSLRightInline{Z. Yin, R. Orozco, and F. J. Herrmann, {``WISER:
Multimodal variational inference for full-waveform inversion without
dimensionality reduction,''} \emph{Geophysics}, vol. 90, no. 2, pp.
A1--A7, 2025, doi: \url{https://doi.org/10.1190/geo2024-0483.1}.}

\bibitem[\citeproctext]{ref-orozco2024velocitymodel}
\CSLLeftMargin{{[}8{]} }%
\CSLRightInline{R. Orozco, H. T. Erdinc, Y. Zeng, M. Louboutin, and F.
J. Herrmann, {``Machine learning-enabled velocity model building with
uncertainty quantification.''} 2024. Available:
\url{https://arxiv.org/abs/2411.06651}}

\bibitem[\citeproctext]{ref-cigdeep2021}
\CSLLeftMargin{{[}9{]} }%
\CSLRightInline{Z. Geng, Z. Zhao, Y. Shi, X. Wu, S. Fomel, and M. Sen,
{``Deep learning for velocity model building with common-image gather
volumes,''} \emph{Geophysical Journal International}, vol. 228, no. 2,
pp. 1054--1070, Sep. 2021, doi:
\href{https://doi.org/10.1093/gji/ggab385}{10.1093/gji/ggab385}.}

\bibitem[\citeproctext]{ref-wang2024controllable}
\CSLLeftMargin{{[}10{]} }%
\CSLRightInline{F. Wang, X. Huang, and T. Alkhalifah, {``Controllable
seismic velocity synthesis using generative diffusion models,''}
\emph{Journal of Geophysical Research: Machine Learning and
Computation}, vol. 1, no. 3, p. e2024JH000153, 2024.}

\bibitem[\citeproctext]{ref-muller2023deep}
\CSLLeftMargin{{[}11{]} }%
\CSLRightInline{A. P. Muller \emph{et al.}, {``Deep-tomography:
Iterative velocity model building with deep learning,''}
\emph{Geophysical Journal International}, vol. 232, no. 2, pp. 975--989,
2023.}

\bibitem[\citeproctext]{ref-Zhao2024}
\CSLLeftMargin{{[}12{]} }%
\CSLRightInline{X. Zhao and A. Curtis, {``Variational prior replacement
in bayesian inference and inversion,''} \emph{Geophysical Journal
International}, vol. 239, no. 2, pp. 1236--1256, Sep. 2024, doi:
\href{https://doi.org/10.1093/gji/ggae334}{10.1093/gji/ggae334}.}

\bibitem[\citeproctext]{ref-deans2002maximally}
\CSLLeftMargin{{[}13{]} }%
\CSLRightInline{M. C. Deans, {``Maximally informative statistics for
localization and mapping,''} in \emph{Proceedings 2002 IEEE
international conference on robotics and automation (cat. No.
02CH37292)}, IEEE, 2002, pp. 1824--1829.}

\bibitem[\citeproctext]{ref-orozco2024aspire}
\CSLLeftMargin{{[}14{]} }%
\CSLRightInline{R. Orozco, A. Siahkoohi, M. Louboutin, and F. Herrmann,
{``ASPIRE: Iterative amortized posterior inference for bayesian inverse
problems,''} \emph{Inverse Problems}, 2024.}

\bibitem[\citeproctext]{ref-psvi}
\CSLLeftMargin{{[}15{]} }%
\CSLRightInline{X. Zhao and A. Curtis, {``Physically structured
variational inference for bayesian full waveform inversion,''}
\emph{Journal of Geophysical Research: Solid Earth}, vol. 129, no. 11,
p. e2024JB029557, 2024.}

\bibitem[\citeproctext]{ref-sbi}
\CSLLeftMargin{{[}16{]} }%
\CSLRightInline{K. Cranmer, J. Brehmer, and G. Louppe, {``The frontier
of simulation-based inference,''} \emph{Proceedings of the National
Academy of Sciences}, vol. 117, no. 48, pp. 30055--30062, 2020, doi:
\href{https://doi.org/10.1073/pnas.1912789117}{10.1073/pnas.1912789117}.}

\bibitem[\citeproctext]{ref-karras2022elucidating}
\CSLLeftMargin{{[}17{]} }%
\CSLRightInline{T. Karras, M. Aittala, T. Aila, and S. Laine,
{``Elucidating the design space of diffusion-based generative models,''}
\emph{Advances in neural information processing systems}, vol. 35, pp.
26565--26577, 2022.}

\bibitem[\citeproctext]{ref-power_scaling}
\CSLLeftMargin{{[}18{]} }%
\CSLRightInline{N. Kallioinen, T. Paananen, P.-C. Bürkner, and A.
Vehtari, {``Detecting and diagnosing prior and likelihood sensitivity
with power-scaling,''} \emph{Statistics and Computing}, vol. 34, no. 1,
p. 57, 2024.}

\bibitem[\citeproctext]{ref-cfg}
\CSLLeftMargin{{[}19{]} }%
\CSLRightInline{J. Ho and T. Salimans, {``Classifier-free diffusion
guidance,''} \emph{arXiv preprint arXiv:2207.12598}, 2022.}

\bibitem[\citeproctext]{ref-predictor_corrector}
\CSLLeftMargin{{[}20{]} }%
\CSLRightInline{A. Bradley and P. Nakkiran, {``Classifier-free guidance
is a predictor-corrector,''} \emph{arXiv preprint arXiv:2408.09000},
2024.}

\bibitem[\citeproctext]{ref-du2023reduce}
\CSLLeftMargin{{[}21{]} }%
\CSLRightInline{Y. Du \emph{et al.}, {``Reduce, reuse, recycle:
Compositional generation with energy-based diffusion models and mcmc,''}
in \emph{International conference on machine learning}, PMLR, 2023, pp.
8489--8510.}

\bibitem[\citeproctext]{ref-BG}
\CSLLeftMargin{{[}22{]} }%
\CSLRightInline{C. E. Jones, J. A. Edgar, J. I. Selvage, and H. Crook,
{``Building complex synthetic models to evaluate acquisition geometries
and velocity inversion technologies,''} \emph{In 74th EAGE Conference
and Exhibition Incorporating EUROPEC 2012}, pp. cp--293, 2012, doi:
\url{https://doi.org/10.3997/2214-4609.20148575}.}

\bibitem[\citeproctext]{ref-judi}
\CSLLeftMargin{{[}23{]} }%
\CSLRightInline{M. Louboutin \emph{et al.}, \emph{Slimgroup/JUDI.jl:
v3.2.3}. (Mar. 2023). Zenodo. doi:
\href{https://doi.org/10.5281/zenodo.7785440}{10.5281/zenodo.7785440}.}

\end{CSLReferences}
\end{document}